# Coronary Calcium Detection using 3D Attention Identical Dual Deep Network Based on Weakly Supervised Learning

Yuankai Huo[1]*, James G. Terry[2], Jiachen Wang[1], Vishwesh Nath[1], Camilo Bermudez[5], Shunxing Bao[1], Prasanna Parvathaneni[1], J. Jeffery Carr[2,3,4] and Bennett A. Landman[1,2,5,6]

[1] Department of Electrical Engineering and Computer Science, Vanderbilt University, Nashville, USA
[2] Departments of Radiology and Radiological Sciences, Vanderbilt University Medical Center, Nashville, USA
[3] Department of Biomedical Informatics, Vanderbilt University Medical Center, Nashville, USA
[4] Department of Cardiovascular Medicine, Vanderbilt University Medical Center, Nashville, USA
[5] Department of Biomedical Engineering, Vanderbilt University, Nashville, USA
[6] Institute of Imaging Science, Vanderbilt University, Nashville, USA

## ABSTRACT

Coronary artery calcium (CAC) is biomarker of advanced subclinical coronary artery disease and predicts myocardial infarction and death prior to age 60 years. The slice-wise manual delineation has been regarded as the gold standard of coronary calcium detection. However, manual efforts are time and resource consuming and even impracticable to be applied on large-scale cohorts. In this paper, we propose the attention identical dual network (AID-Net) to perform CAC detection using scan-rescan longitudinal non-contrast CT scans with weakly supervised attention by only using per scan level labels. To leverage the performance, 3D attention mechanisms were integrated into the AID-Net to provide complementary information for classification tasks. Moreover, the 3D Gradient-weighted Class Activation Mapping (Grad-CAM) was also proposed at the testing stage to interpret the behaviors of the deep neural network. 5075 non-contrast chest CT scans were used as training, validation and testing datasets. Baseline performance was assessed on the same cohort. From the results, the proposed AID-Net achieved the superior performance on classification accuracy (0.9272) and AUC (0.9627).

## 1. INTRODUCTION

Coronary artery calcium (CAC) indicates the presence of advance coronary plaque and is a strong and independent predictor of myocardial infarction and cardiovascular death prior to age 60 years [2]. In typical practice, non-contrast computed tomography (CT) scans are obtained and then manually annotated for calcified lesions in each of the coronary arteries. The manual detection with quality control has been regarded as the standard assessment for CAC measurements. However, slice-wise manual tracing is time and resource consuming, and impracticable for a large population. Therefore, it is appealing to develop automatic CAC measurement methods to alleviate the manual efforts. In the past decade, many automated CAC detection approaches have been developed to achieve CAC detection and estimations from non-contrast CT [3, 4]. In recent years, deep convolutional neural network (DCNN) based methods have shown their superior performance on many medical image analysis tasks in terms of accuracy and efficiency. For DCNN based CAC detection, Lessmann et al. [5] proposed a DCNN to detect CAC using hierarchical structures with two DCNN networks. Santini et al. [6] proposed a patch-based CAC segmentation method. Shadmi et al. [7] introduced the fully convolutional network (FCN) for CAC segmentation. However, the previous efforts were typically trained in 2D or 2.5D fashion rather than 3D. Moreover, voxel/lesion level manual annotation were typically required to train a DCNN based CAC detection in previous studies. Meanwhile, recent studies [8, 9] trains a deep network using scan-rescans data, which has shown the advantages on Diffusion Weighted magnetic resonance imaging (MRI).

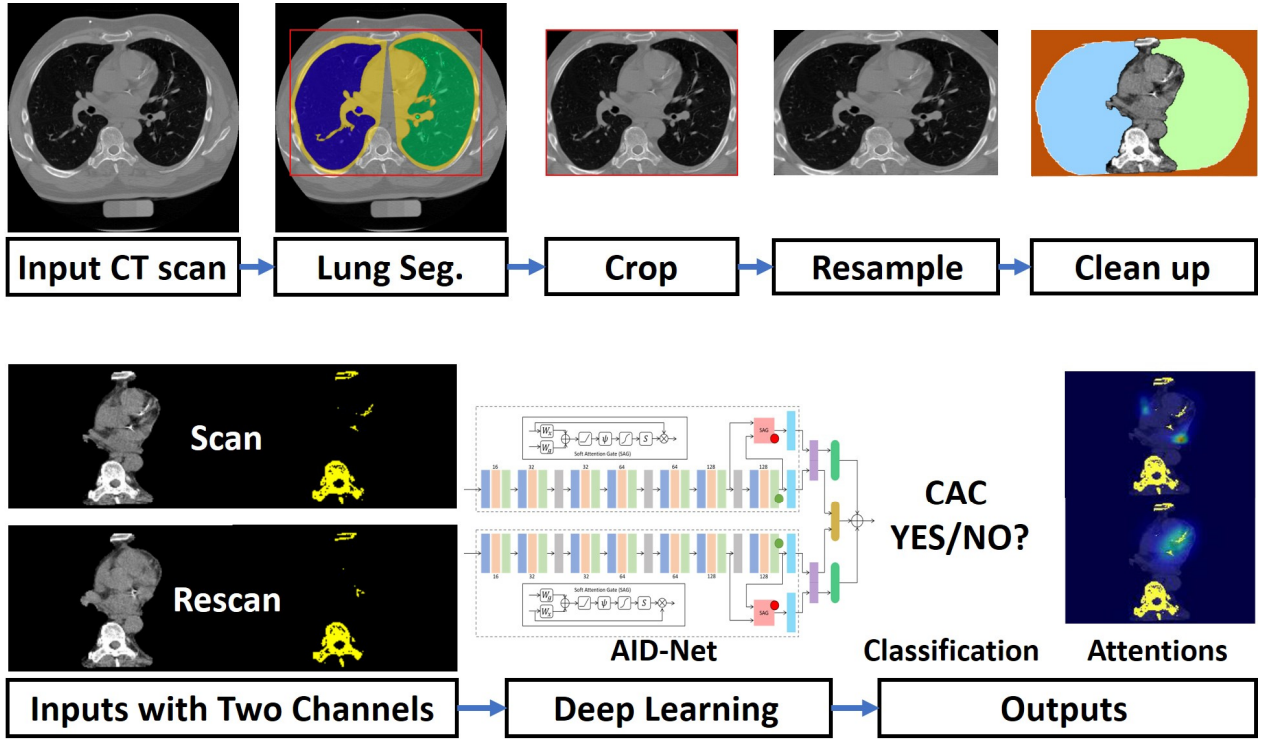

Figure 1. The entire processing pipeline of AID-Net method. The upper panel shows the preprocessing steps, while the lower panel shows the deep learning procedures.

In this paper, we propose a 3D DCNN based CAC detection method, called attention identical dual network (AID-Net), to perform CAC detection using weakly supervised attentions on per-subject level labels. Specifically, the purpose of the proposed AID-Net is to determine if a non-contrast CT volume has CAC or not. The proposed method has three features: (1) 3D identical dual network is proposed to use longitudinal CT scans for CAC detection, (2) a 3D training attention gate is integrated to the network to leverage the performance, (3) 3D weakly supervised attention mechanisms were introduced to explain the behavior of the proposed network (visualize the location) only using subject level weak labels. To train the network, we used 5075 non-contrast chest CT scans with Agatston calcium score (Agatston score) [10] without using the voxel/lesion level manual annotations.

## 2. METHOD

**2.1 Preprocessing**

The same automatic preprocessing procedures (Figure 1) are deployed on each scan to normalize the data. Briefly, a whole lung segmentation method [11] was first applied to the chest CT scan (https://github.com/lfz/DSB2017). Then, the 3D sub-space containing two whole lung regions are cropped (contains the heart) as the useable region. On such sub-space, the Hounsfield unit (HU) intensity within the lung segmentation region and the intensities outside the lung convex hull mask were set to -200 HU. For the remaining voxels, HU scores are windowed to -200 for HU<-200 and +600 for HU>600. Meanwhile, another 3D mask is obtained by setting the threshold HU>130 to be used as another input channel for AID-Net.

**2.2 Attention Identical Dual Network (AID-Net)**

The previously mentioned 3D cropped CT volume and mask volume are used as two input channels for AID-Net. The identical dual network, inspired by Siamese network [12], is introduced to take advantage of longitudinal training scans. To leverage the classification performance, we extend a recent proposed soft attention gate (SAG) [1] from 2D to 3D as a training attention to provide complementary local features, which are concatenated with the major global averaging pooling features for final dense layer. The total loss function of the AID-Net consists of three terms:

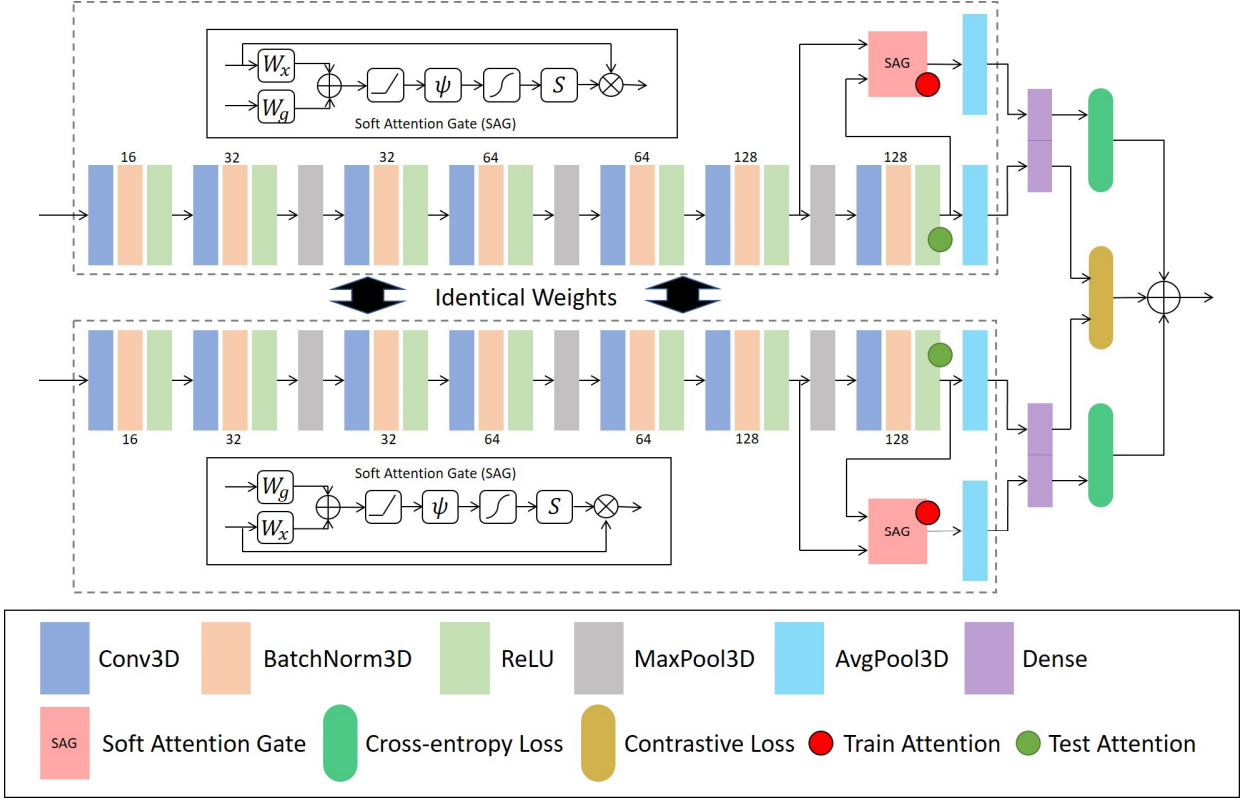

Figure 2. The network structure of proposed AID-Net. The inputs of the network are the longitudinal scan and rescan data, where each data consists of two channels: preprocessed CT and HU>130 mask. The outputs of the network are the classification results and training and testing attention maps. The definition on symbols of SAG subnetwork is the same as [1].

$$Loss_{total} = Loss_{cross-entropy1} + Loss_{cross-entropy2} + \lambda \cdot Loss_{contrastive}$$

where the $Loss_{cross-entropy1}$ and $Loss_{cross-entropy2}$ are the cross-entropy loss for the two paths in AID-Net. The $Loss_{contrastive}$ is the contrastive loss [13], which measures the similarities between the outputs of two paths as

$$Loss_{contrastive} = (1-Y)\frac{1}{2}(D_w)^2 + (Y)\frac{1}{2}(\max(0, m - D_w))^2$$

where $D_w$ is the Euclidean distance between the outputs of two paths in AID-Net. $m$ is a coefficient to limit the range of maximization. $Y = 0$ if the inputs are deemed similar, and $Y = 1$ if they are deemed dissimilar. $\lambda$ is an constant coefficient to decide the weight of contrastive loss.

### 2.3 Training and Testing Attention

We proposed 3D SAG as the training attention, which extended the 2D based SAG into 3D fashion. To be compatible with large memory consumption in 3D, the channel number of the intermediate layer was set to 32. For 3D SAG, the feature maps before the last global average pooling layer were used as gate ($g$) to filter the feature maps from higher level layers, which worked as another complimentary inputs for classification. The attention map from the lower level layers typically yields a larger reception field. Therefore, the SAG enables the AID-Net to consider higher level and some task related local features when making the classification decision. If the 3D SAG is excluded in the network, the AID-Net becomes ID-Net without training attention.

To characterize the attention mechanism of the main stream in AID-Net, the Grad-CAM [14] method has been extended from 2D to 3D in this work to visualize the attention maps, which were related to the classification tasks. Using the 3D Grad-CAM, the attention maps were able to explain the behavior of the DCNN and localize the CAC. The Grad-CAM attention was generated before the global average pooling layer as shown in Figure 2.

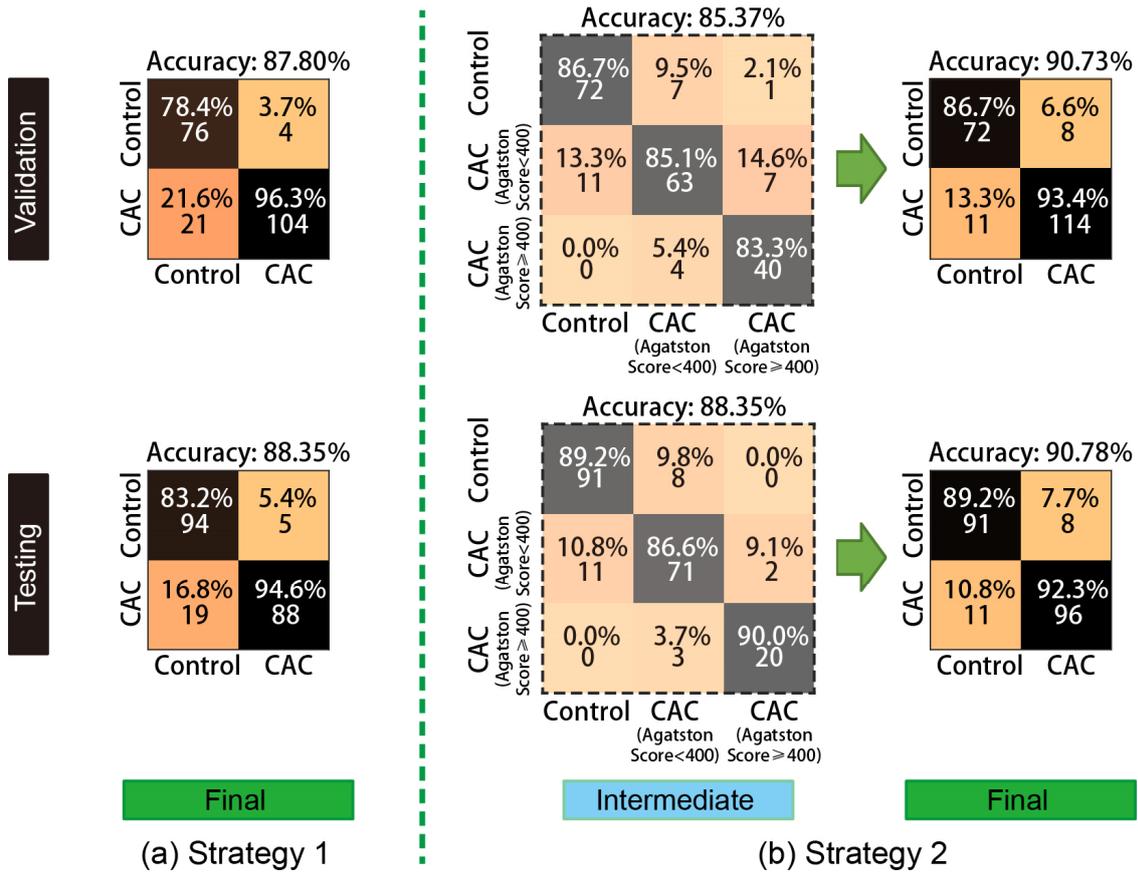

Figure 3. Two strategies of training a CAC classification network. The "Strategy 1" is the traditional design for a two-class classification task. To further leverage the performance, we split the CAC to two sub-categories: (Agatston score < 400 and Agatston score ≥ 400). Then, the 3×3 classification matrix for prediction were converted to two class problem for better prediction performance.

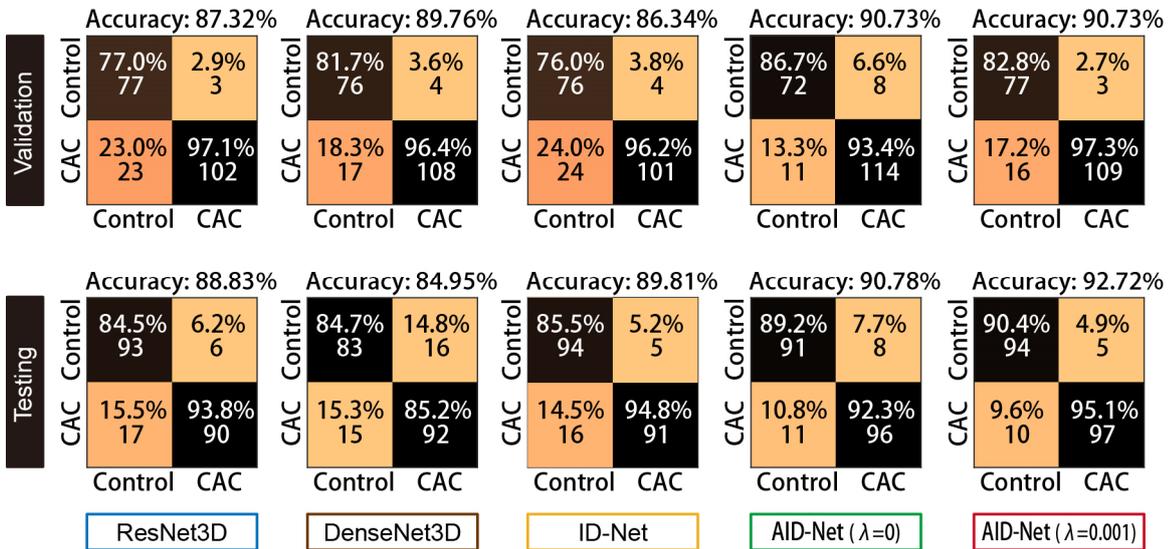

Figure 4. The final CAC classification results are showed as 2×2 classification metrices.

## 2.4 Experiments

To evaluate the classification performance of proposed AID-Net, the 3D ResNet [15] with 101 layers as well as the 3D DenseNet [16] with 121 layers were employed as the baseline methods. All 4664 scans from 2332 subjects were used as inputs for baseline methods, while the 2332 scan-rescan pairs were used as inputs of the AID-Net. To evaluate the performance of 3D SAG training attention, the ID-Net was also evaluated. For AID-Net, two loss coefficients ($\lambda=0$ and $\lambda=0.001$) were compared. After training the network, 205 scans were used as validation data and another set of 206 scans were used as the independent testing data. The parameters and epochs with the best performance were tuned through 205 validation scans and been applied to 206 withheld testing scans. For each scan, the clinical measure of CAC, the Agatston score is available. To form the CAC detection task as a classification problem, we evaluated two strategies. Presence verses absence of CAC using a threshold of any score > 0. The second strategy used a three-class approach with scores of 0, 1-399 and 400 and greater. We did not further split the data with more sub-categories using other thresholds (e.g., Agatston score at 100, 200, 300, etc.) since the target of the proposed work is to detection "Yes/No" for CAC rather than Agatston score regression. Moreover, the Agatston score $\geq 400$ is typically regarded as severe CAC scenarios. As a result, 1107 control scans, 848 scans with Agatston score < 400, 377 scans with Agatston score $\geq 400$ were used in the training procedures. As shown in Figure 3, the latter strategy yields better performance than categorizing all CAC patients as one class for deep learning.

The network was trained using the following parameters: learning rate = 0.0001, maximum epoch number = 100, input resolution = 192×128×64, batch size = 2, $\lambda$ = 0.001, optimizer = Adam. The network was implemented using PyTorch = 0.4 and CUDA = 8.0. The experiments were performed on NVIDIA GeForce Titan GPU with 12GB memory.

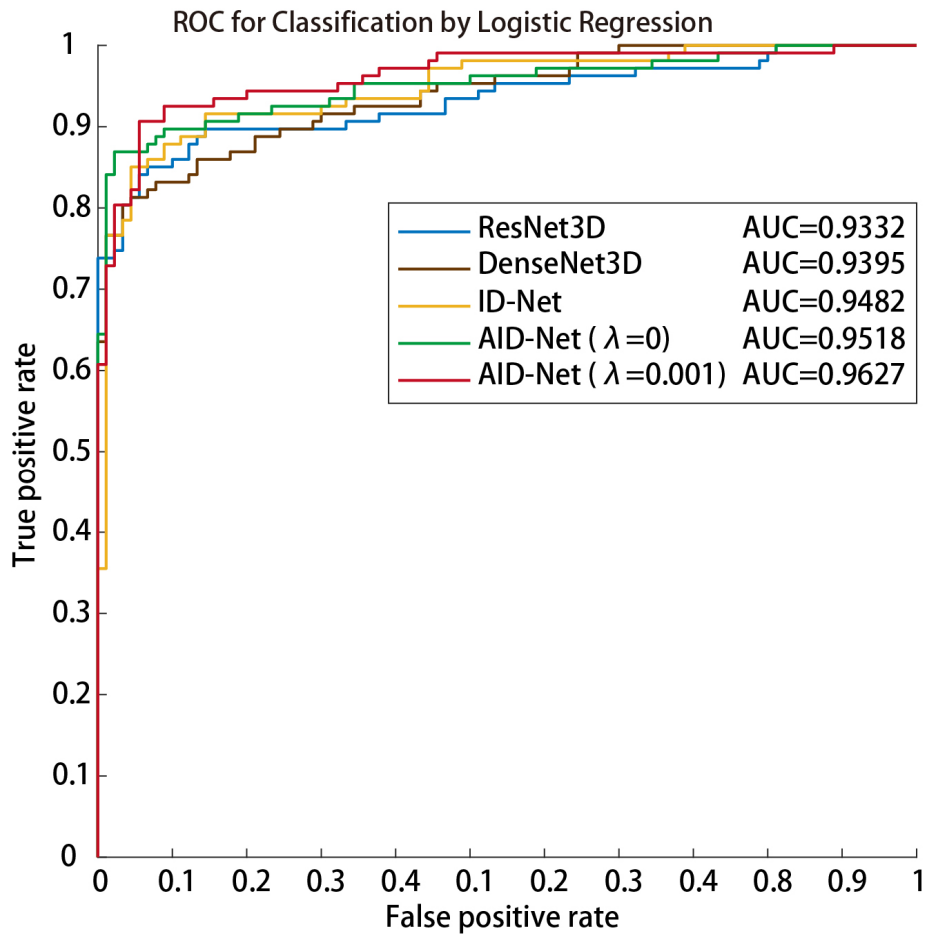

Figure 5. The Receiver Operating Characteristic (ROC) curves and Area under the ROC Curve (AUC) are presented.

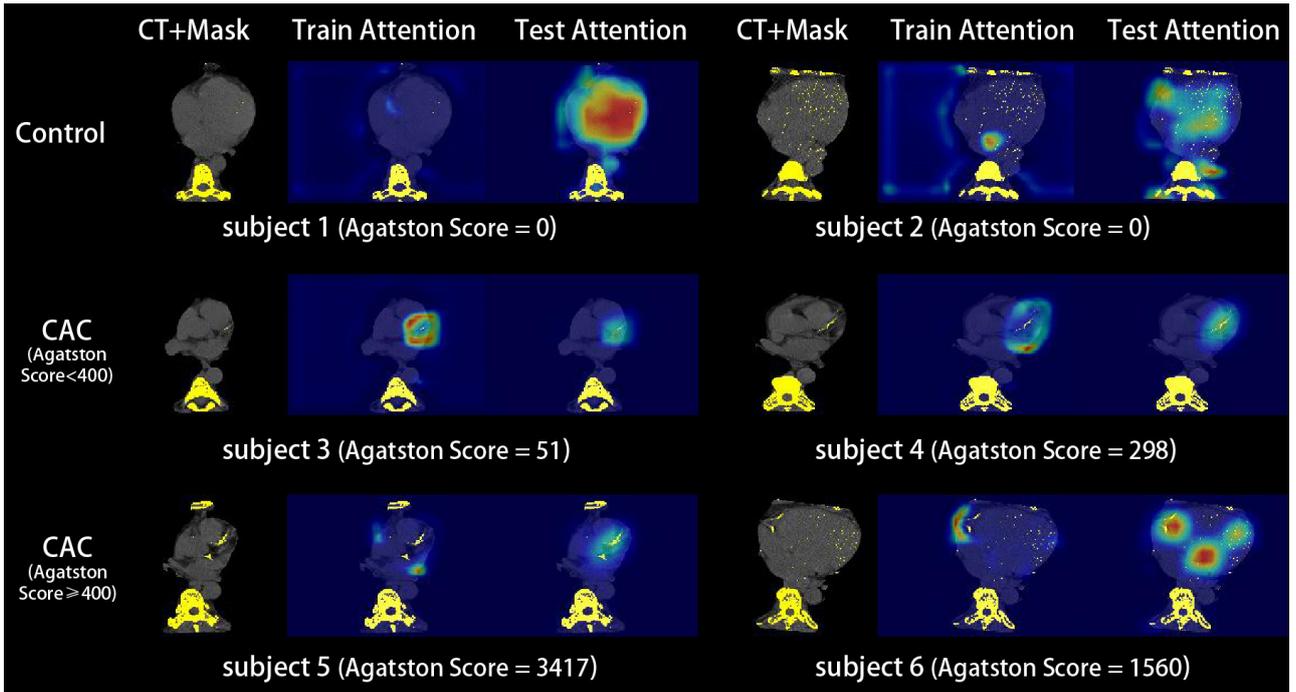

Figure 6. The training and testing attention mechanisms were presented in this figure. The 3D SAG training attention typically achieves local features from higher level layers, which provides the complementary information for classification decision. The 3D Grad-CAM testing attention reflects the behavior of the deep networks when making classification decisions. The Grad-CAM attention maps are also be able to roughly localize the CAC.

## 3. RESULTS

Figure 4 shows the 2×2 classification matrices of different methods on CAC detection, which is formed as classification problem. Figure 5 shows the Receiver Operating Characteristic (ROC) and Area Under the ROC (AUC) curves for different methods. From such results, the proposed AID-Net with $\lambda = 0.001$ achieved the best overall classification performance. Figure 6 presents the 3D training attention gate from SAG as well as the 3D testing attention heatmaps form 3D Grad-CAM. From the results, the training attention provides the complementary attentions on the task related local features for better prediction performance (from red circle in Figure 2). The 3D Grad-CAM (testing attention) maps show that the testing attention is able to localize the CAC location using weakly annotated attention mechanism (from green circle in Figure 2). Subject 1 and 2 in Figure 6 are two representative cases for Controls. For subject 2, although a large number of voxels are marked as HU>130, the proposed AID-Net is able to distinguish them from CAC by considering the spatial locations of such voxels. For subject 3, even though a small number of voxels are marked as HU>130, the proposed AID-Net is able to locate the CAC. Moreover, the training attention focused on the local features around the CAC, while the testing attention located the CAC. The similar situations were presented for subject 4 to 6.

## 4. DISCUSSION

In this paper, we present the AID-Net, which is able to perform CAC detection with accuracy = 0.9272, and AUC = 0.9627. The performance of the proposed AID-Net achieved superior performance compared with ResNet, DenseNet, and ID-Net. The attention mechanisms were presented as qualitative results in this paper without quantitative validation due the lack of the per voxel/lesion annotation. In the next phrase, we would like to validate the attention model using the per voxel/lesion annotation. The performance of the proposed method would be further improved when such annotations are available. Moreover, the detailed heart structures can be achieved by DCNN-based 3D medical image segmentation techniques [17, 18], which is able to be combined with CAC detection, domain adaptation [19, 20] as a multi-task learning procedure [21].

Prior efforts on automatic CAC detection methods would yield better performance on the testing cohort. However, the previous works typically required resource intensive per voxel/lesion annotations, which were not required in the proposed AID-Net. To the best of our knowledge, this work is the first deep learning method on CAC detection using weakly supervised attention. As the per voxel/lesion annotations are not require, the proposed method could be trained on larger clinical cohorts with Agatston scores available.

## ACKNOWLEDGEMENT

This research was supported by NSF CAREER 1452485, NIH grants 5R21EY024036, 1R21NS064534, 1R01EB017230 (Landman), and 1R03EB012461 (Landman). This study was in part using the resources of the Advanced Computing Center for Research and Education (ACCRE) at Vanderbilt University, Nashville, TN. This project was supported in part by ViSE/VICTR VR3029 and the National Center for Research Resources, Grant UL1 RR024975-01, and is now at the National Center for Advancing Translational Sciences, Grant 2 UL1 TR000445-06. We gratefully acknowledge the support of NVIDIA Corporation with the donation of the Titan X Pascal GPU used for this research. The imaging dataset(s) used for the analysis described were obtained from ImageVU, a research resource supported by the VICTR CTSA award (ULTR000445 from NCATS/NIH), Vanderbilt University Medical Center institutional funding and Patient-Centered Outcomes Research Institute (PCORI; contract CDRN-1306-04869).